\documentclass[conference]{IEEEtran}
\IEEEoverridecommandlockouts
\usepackage{cite}
\usepackage{amsmath,amssymb,amsfonts}
\usepackage{textcomp}

\usepackage{algorithm}
\usepackage{algorithmic}
\usepackage{graphicx} 
\usepackage{array} 
\usepackage{multirow} 
\usepackage{caption} 
\usepackage{amsmath} 
\usepackage{booktabs} 
\usepackage{enumitem}
\usepackage{amsmath}  
\usepackage{amssymb}  
\usepackage{array}    
\usepackage{colortbl} 
\usepackage{subcaption}

\def\BibTeX{{\rm B\kern-.05em{\sc i\kern-.025em b}\kern-.08em
    T\kern-.1667em\lower.7ex\hbox{E}\kern-.125emX}}
\begin{document}

\title{CASA: Class-Agnostic Shared Attributes in Vision-Language Models for Efficient Incremental Object Detection}

\author{
    Mingyi Guo\textsuperscript{\rm 1, *},
    Yuyang Liu\textsuperscript{\rm 1, *, \dag},
    Zhiyuan Yan\textsuperscript{\rm 1},    
    Zongying Lin\textsuperscript{\rm 1},
    Peixi Peng\textsuperscript{\rm 1,2},
    Yonghong Tian\textsuperscript{\rm 1,2,3, \dag} \\
\textsuperscript{\rm 1}School of Electronic and Computer Engineering, Peking University,  
\\
\textsuperscript{\rm 2}Peng Cheng Laboratory, 
\textsuperscript{\rm 3}School of Computer Science, Peking University
\\
{\tt\small myguo@stu.pku.edu.cn, liuyuyang13@pku.edu.cn, yhtian@pku.edu.cn}
}
\maketitle

\begin{abstract}
Incremental object detection is fundamentally challenged by catastrophic forgetting. A major factor contributing to this issue is \textit{background shift}, where background categories in sequential tasks may overlap with either previously learned or future unseen classes.
  To address this, we propose a novel method called Class-Agnostic Shared Attribute Base (CASA) that encourages the model to learn category-agnostic attributes shared across incremental classes.
  Our approach leverages an LLM to generate candidate textual attributes, selects the most relevant ones based on the current training data, and records their importance in an assignment matrix. For subsequent tasks, the retained attributes are frozen, and new attributes are selected from the remaining candidates, ensuring both knowledge retention and adaptability.
Extensive experiments on the COCO dataset demonstrate the state-of-the-art performance of our method. 
\begingroup
\renewcommand{\thefootnote}{}
\footnotetext{Supported by Shenzhen Peacock Team KQTD 20240729102051063 and the China Postdoctoral Science Foundation under Grant Number BX20240013 and 2024M760113. 

Equal Contribution: * ; Corresponding author: \dag}
\endgroup
\end{abstract}

\begin{IEEEkeywords}
incremental object detection learning, vision-language models, efficient learning
\end{IEEEkeywords}

\section{Introduction}
\label{sec:intro}
Object detection models have achieved remarkable success in recognizing and localizing objects within a fixed set of categories. They assume a static training environment where all object categories are known and labeled beforehand. In real-world scenarios, as new object categories emerge over time, the need for continuous updates to these models becomes crucial—a process known as incremental object detection (IOD)~\cite{feng2022overcoming,liu2023augmented}. However, IOD also suffers significant challenges, particularly in managing \textit{background drift}~\cite{cermelli2020modeling}. \textit{Background drift} occurs when objects belonging to previous or future tasks in IOD are not annotated in the current task and are instead assigned to the background class. This can cause serious misclassification issues~\cite{liu2023augmented}, as the model may later confuse these background objects with newly introduced categories or fail to recognize them altogether. This highlights a key challenge in IOD: how to maintain the ability to generalize across tasks without comprehensive annotations.

In traditional object detection, accurate classification relies on extracting and utilizing shared visual features, \textit{i.e.,} shape, texture, and color. These fine-grained semantic details are crucial for distinguishing objects from the background, particularly as new categories appear. Research in transfer learning and domain adaptation highlights that effective generalization depends on the ability to leverage these common attributes. However, in IOD, the evolving background class with each new task complicates the traditional models struggle to consistently capture and apply these shared features.

Recent advances in vision-language models~\cite{radford2021learning,zhou2022conditional} offer promising solutions to these challenges. By integrating visual and textual data, these models provide a richer contextual understanding of both objects and their backgrounds. This cross-modal approach enhances the ability to retain and align shared semantic information across tasks, thereby mitigating the effects of \textit{background drift} and improving the robustness of IOD. The inability to maintain a coherent representation of these semantic relationships across tasks is thus a fundamental problem in IOD, compromising its overall performance as it encounters new categories. 

To tackle these challenges, we propose Class-Agnostic Shared Attributes (CASA), a novel approach that leverages vision-language foundation models to address \textit{background drift}. It captures and utilizes common semantic information across incremental classes. By employing LLM, we generate candidate textual attributes relevant to the object categories, curate these attributes based on their relevance to the current training data, and record their importance in an attribute assignment matrix. For subsequent tasks, we freeze the retained attributes while continuing to select and update relevant attributes, ensuring that the model adapts incrementally without losing previously learned knowledge. Building on OWL-ViT~\cite{minderer2022simple}, CASA achieves only a minimal increase in parameter storage (0.7\%) through parameter-efficient fine-tuning, significantly enhancing its scalability and adaptability. Experiments on the COCO dataset demonstrate its effectiveness, achieving state-of-the-art performance across both two-phase and multi-phase incremental learning scenarios.

In summary, our contributions are as follows: 
\begin{itemize}
    \item We propose Class-Agnostic Shared Attributes (CASA) for leveraging common semantic information across categories in IOD, overcome the \textit{background drift}. 
    \item Our method utilizes a frozen vision-language foundation model with parameter-efficient fine-tuning that only increases parameter storage by 0.7\%, significantly improving the scalability and adaptability of IOD.
    \item Extensive two-phase and multi-phase experiments on COCO dataset demonstrate the effectiveness and efficiency of CASA, achieving SOTA performance in IOD.
\end{itemize}

\section{Related Works}
\label{sec:rw}

\subsection{Vision-Language Models}
Vision-language models have emerged as powerful tools for understanding and integrating visual and textual data, enabling more comprehensive and context-aware representations of objects and scenes. One of the most prominent models is CLIP (Contrastive Language-Image Pretraining)~\cite{radford2021learning}, which leverages large-scale image-text pairs to learn a joint embedding space where visual and textual modalities are aligned. This approach has inspired subsequent research in vision-language models, such as ALIGN and Florence, which further enhance the ability to generalize across diverse datasets and tasks. Recent advancements in vision-language models about object detection and classification have significantly improved robustness and adaptability, particularly in scenarios involving novel or evolving categories~\cite{radford2021learning}. Among these methods, OWL-ViT~\cite{minderer2022simple}, an open-world learning vision transformer, stands out by enabling more flexible and scalable detection and classification. FOMO~\cite{zohar2023open} based on OWL-ViT utilizes the attributes of known objects to recognize unknown objects in the open-world environment.
\subsection{Incremental Learning with Pre-trained Models (PTMs)}
Traditional incremental learning methods~\cite{EWC,icarl,DEN,liu2021l3doc} usually start with a model trained from scratch. With the emergence of a variety of foundation models, incremental learning with PTMs aims to leverag their strong generalization to downstream tasks.
It can be divided into three strategies: Prompt-based, Representation-based, and Model Mixture-based methods~\cite{zhou2024continual}. Prompt-based methods leverage the strong generalization capabilities of PTMs by using prompts to perform lightweight updates without fully fine-tuning all model parameters~\cite{jung2023generating}. Representation-based methods directly utilize the generalization capabilities of PTMs to construct classifiers without making significant adjustments to the model itself~\cite{zhou2024expandable}. Model Mixture-based methods design a set of models during the learning process and employ techniques like model merging and ensemble learning to make final predictions~\cite{zheng2023preventing,yang-etal-2025-parameter}. By combining different models, these methods aim to capitalize on the strengths of each model to enhance overall learning performance.
\subsection{Incremental Object Detection}
Incremental Object Detection~\cite{shmelkov2017incremental,li2017learning,icarl,liu2023continual,kim2024sddgr} presents unique challenges compared to standard object detection, particularly in managing the evolving nature of the background class and ensuring that newly introduced categories do not interfere with previously learned ones.
Existing IOD methods can be broadly classified into three categories: Knowledge Distillation (\textit{e.g.,} LWF~\cite{li2017learning}), Replay(\textit{e.g.,} iCaRL~\cite{icarl}, ABR~\cite{liu2023augmented})  and Regularization methods(\textit{e.g.,} ERD~\cite{feng2022overcoming}, LID~\cite{yanzhiyuan}).
In recent years, with the popularity of the Transformer architecture, more and more methods have tried to use DETR as the baseline. 
CL-DETR~\cite{liu2023continual} extends the DETR architecture to support incremental learning scenarios by leveraging transformer-based representations.
Furthermore, CIOD~\cite{kim2024sddgr} specifically focuses on maintaining detection accuracy across a growing number of classes by integrating adaptive feature extractors and regularization strategies.
Additionally, the integration of vision-language models into IOD is a promising direction. These methods demonstrate that textual prompts can provide more semantic information for the representation of visual modality.
\begin{figure*}[h]
    \centering
    \includegraphics[width=0.95\textwidth]{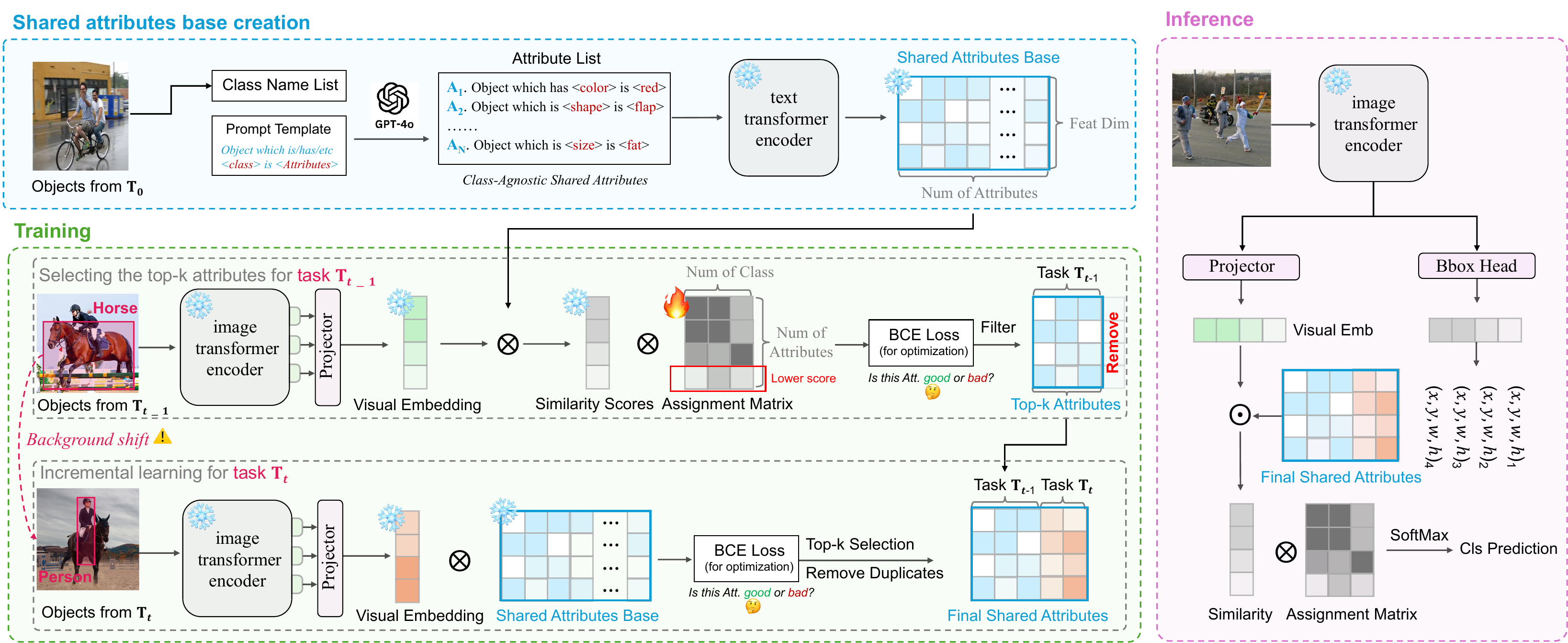}
    \caption{Illustration of our proposed \textbf{C}lass-\textbf{A}gnostic \textbf{S}hared \textbf{A}ttribute (CASA). We leverage LLMs to generate the shared attribute base $E_a$ and then select the most relevant ones $\hat{E}_a^t$ based on the current training data, documenting their significance in an attribute assignment matrix $A^t$. 
    In subsequent tasks, we retain and freeze these selected attributes, continuing the process by choosing from the remaining candidates and appending them after $\hat{E}_a^{t-1}$, and updating the attribute assignment matrix.}
    \label{fig:1}
    \vskip -0.1in
\end{figure*}

\section{Method}

 \subsection{Problem Formulation and Overview}
IOD is an extension topic of conventional object detection, allowing for learning new categories without forgetting old ones. 
Specifically, when the model learns a new class during Task $T_t$, it should retain its ability to recognize the old classes ranging from Task $T_0$ to Task $T_{t-1}$. 
In this work, we propose encouraging the model to learn category-agnostic shared attributes across different objects, rather than overfitting to class-specific features, which can result in catastrophic forgetting.
The pipeline of our method is shown in Figure~\ref{fig:1}.

\subsection{Shared Attributes Base Creation}
In this module, we aim to generate the class-agnostic attributes base using the data from the first-given task ($T_0$).
Specifically, we first generate a large amount of attributes corresponding to different objects using LLMs (\textit{i.e.,} Gpt-4o). 
Then we apply the attribute prompt template to obtain $N$ pieces of attribute texts, such as ``\textit{object which has color is red.}'' 
Note that the used prompts are designed to be class-agnostic, where we use abstract ``object" to replace the specific name of the object.
This manner encourages the model to learn more common and shared attributes across different objects.
Subsequently, we compress these texts with a Text Encoder~\cite{radford2021learning} to obtain the attribute embedding base, denoted as $E_a = [\mathbf{e}^t_{A_1}, \mathbf{e}^t_{A_2}, ... , \mathbf{e}^t_{A_N}]\in\mathbb{R}^{D\times N}$. 
The created attribute embedding base $E_a$ will be frozen and used for all tasks. 
Note that not all attributes are informative and general for our task.

\subsection{Training}

\subsubsection{Motivation}
The primary challenge in IOD is background shift, where the model tends to overfit to object-specific visual features from previous tasks, limiting its ability to retain prior knowledge and adapt to new, unseen classes.

To address this challenge, we propose leveraging vision-language models to learn a class-agnostic shared attributes base, which can be utilized for more general and robust detection. This approach involves solving two core problems: (a) \textit{How can we identify the most informative and general attributes from the shared attribute base?} (b) \textit{How can we effectively utilize these selected attributes for object detection?} The following sections provide detailed technical methodologies.

\subsubsection{Attributes Selection and Attributes Assignment}
To address the first problem, we identify the most informative and general attributes by matching attributes embeddings with the visual embeddings ($\mathbf{e}^v$) extracted from the visual encoder. Specifically, during Task $T_t$, given the attributes embedding $\mathbf{e}^t_A$ and the visual embedding $\mathbf{e}^v$, the similarity vector $S^t$ is computed as:
\begin{equation}
    S^t = \text{CosSim}(\mathbf{e}^v, \mathbf{e}^t_A).
    \label{eqn:s_A}
\end{equation}

Here, the similarity vector $S^t$ reflects the relevance of each attribute, with higher values indicating more informative attributes. This step ensures the selection of attributes that are meaningful for object detection.

To map the selected attributes to object categories for detection, we introduce an \textit{assignment matrix}, where each element of the assignment matrix represents the score of an attribute being associated with a specific category. In Task $T_t$, the assignment matrix is defined as $A^t \in \mathbb{R}^{N \times C^{1:t}}$, where $C^{1:t}$ denotes the total number of categories learned up to Task $T_t$, and $C^t$ represents the number of new categories introduced in Task $T_t$. Initially, the assignment matrix is initialized with arbitrary values between 0 and 1.

The similarity vector $S^t$ is then used to update the assignment matrix $A^t$. Specifically, for the newly introduced classes in Task $T_t$, we extract the last $C^t$ columns of the assignment matrix, denoted as $A^{t-1:t}$. The class probability $p_{cls}$ for these new classes is computed as:
\begin{equation}
    p_{cls} = \text{Sigmoid}(A^{t-1:t} S^t).
    \label{eqn:logits}
\end{equation}

Subsequently, the Binary Cross-Entropy (BCE) loss is calculated between the predicted probabilities $p_{cls}$ and the one-hot encoded targets $y$.
Additionally, we add a regularization loss to enforce sparsity. 
The overall loss function \( \mathcal{L}_{upd} \) can be formulated as:
\begin{equation}
    \mathcal{L}_{upd} = \mathcal{L}_{BCE}(p_{cls}, y).) + \lambda \sum_{i,j} |A^{t-1:t}_{t_{i,j}}|,
    \label{eqn:loss}
\end{equation}
where $\lambda$ is a tunable hyperparameter, which is set to 0.01 fixed in our work.
This loss is used to iteratively refine the assignment matrix $A^{t-1:t}$, ensuring that it captures meaningful relationships between attributes and the new categories.

\subsubsection{Attribute Filtering and Incremental Learning}
Once the assignment matrix $A^{t-1:t}$ has been updated, we proceed to filter the shared attributes. Specifically, for each category, we select the top $H_a$ representative attributes, where $H_a$ is a predefined constant. To achieve this, $A^{t-1:t}$ is flattened into a one-dimensional vector, and the top $C^t \times H_a$ elements with the highest scores are selected. The remaining elements are set to zero, and the matrix is reshaped back into a binary matrix, where each element is either 0 or 1.

To implement incremental learning, the assignment matrix from the previous task, $A^{t-1}$, is concatenated with the current matrix $A^{t-1:t}$ to form a new matrix $A^t \in \mathbb{R}^{N \times C^{1:t}}$. This updated matrix is saved for subsequent tasks, enabling seamless incremental learning across multiple tasks.

\begin{table*}[h]
\centering
\caption{CASA results (\%) on COCO 2017 in \emph{two-phase setting} 70+10. The best performance is highlighted in \textbf{bold}.}
\vskip -0.1in
\renewcommand{\arraystretch}{1.0}
\small
\label{tab:70+10}
{
\begin{tabular}{c|l|c|cccccc}
\toprule
{Scenarios} & \multicolumn{1}{c|}{{Method}}  & {Baseline} & $AP$ & {$FPP$} & $AP_{.5}$ & {$FPP$} & $AP_{.75}$ & {$FPP$} \\ \hline
\multirow{1}{*}{80} 
    & Joint Training & OWL-ViT & 42.1 & non & 61.8 & non & 47.1 & non \\ \hline
\multirow{2}{*}{70} 
    & --- & Deformable DETR & 43.4 & non &  62.8 & non & 47.2 & non \\ 
    & --- & OWL-ViT & 43.76 & non & 63.43 & non & 48.37 & non \\ 
    \hline
\multirow{9}{*}{70 + 10} 
     & ERD~\cite{feng2022overcoming} & UP-DETR & 36.2 & ---& 54.8 & ---& 39.3 &--- \\   
     & CL-DETR~\cite{liu2023continual} & UP-DETR & 37.6 &--- & 56.5 &--- & 39.4 & ---\\ 
     & LwF~\cite{li2017learning} & Deformable DETR & 24.5 &--- & 36.6 &--- & 26.7 &--- \\ 
     & iCaRL~\cite{icarl} & Deformable DETR & 35.9 &--- & 52.5 & ---& 39.2 & ---\\ 
     & ERD~\cite{feng2022overcoming} & Deformable DETR & 36.9 & ---& 55.7 & ---& 40.1 &--- \\ 
     & CL-DETR~\cite{liu2023continual} & Deformable DETR & 40.1 & ---& 57.8 & ---& 43.7 &--- \\ 
     & VLM-PL~\cite{kim2024vlm} & Deformable DETR & 39.8 & --- & 58.2 & --- & 43.3 & --- \\
     & CIOD~\cite{kim2024sddgr} & Deformable DETR & 40.9 & 1.9 & 59.5 & 2.2 & 44.8 & 1.8 \\ 
     & CASA & OWL-ViT & \textbf{42.2} &\textbf{ -0.01 }& \textbf{61.0} & \textbf{-0.01} & \textbf{46.6} & \textbf{-0.01} \\ 
\bottomrule
\end{tabular}
}
\vskip -0.1in
\end{table*}

\subsubsection{Attribute Sharing and Class-Agnostic Learning}
During IOD, the assignment matrix $A^t$ allows for attribute sharing across tasks. For any row in $A^t$, a value of 1 indicates that the corresponding attribute is relevant for a specific category, while a value of 0 indicates irrelevance. Rows in $A^t$ where all values are 0 are removed, ensuring that only useful attributes are retained. Additionally, attributes shared across tasks are preserved by maintaining indices from the previous task, denoted as $id_{t-1}$. These indices are updated to form $id_t$ for the current task. Using $id_t$, we filter the attribute embeddings and assignment matrix, retaining only rows corresponding to meaningful indices. This process ensures that shared attributes are utilized effectively, enabling the model to learn in a category-agnostic manner and improving its generalization across tasks.

Besides, there are another two steps written in supplementary materials II. By addressing both attribute selection and utilization, our method effectively mitigates the challenge of background shift and enables robust IOD.

\subsection{Inference}
During inference, the trained model utilizes the category-agnostic shared attributes and the assignment matrix to detect objects incrementally across tasks. Given an input image, the visual encoder extracts the visual embeddings $\mathbf{e}^v$, which are then matched with the attribute embeddings $\mathbf{e}_A^t$ using the similarity computation defined in Equation~\ref{eqn:s_A}. This results in a similarity vector $S^t$, which highlights the relevance of each attribute for the given input.
Using the assignment matrix $A^t$ from the current task, the model maps the similarity vector $S^t$ to the class probabilities $p_{cls}$ for all categories learned up to the current task $T_t$. This mapping is performed as described in Equation~\ref{eqn:logits}, ensuring that the output reflects the contributions of both new and previously learned attributes. The resulting probabilities are used to classify the detected objects into one of the learned categories, while maintaining consistency with the knowledge retained from earlier tasks.

By leveraging the binary structure of the assignment matrix $A^t$, the model dynamically filters irrelevant attributes, focusing only on those that are meaningful for object detection. Additionally, the preserved shared attributes across tasks ensure that the model generalizes well to new categories without forgetting prior knowledge. This process enables the detection of objects from both old and new categories in a seamless and efficient manner, demonstrating the effectiveness of the proposed method in addressing the challenges of IOD.

\section{Experiments}
\label{exp}
\subsection{Dataset and Evaluation Metrics}
We evaluate our method on the MS COCO 2017 dataset, which is widely used in IOD. Via both two-phase and multi-phase setting, we conduct a comprehensive comparison with other IOD methods in terms of evaluation metrics $AP$, $AP_{.5}$ and $AP_{.75}$. We also evaluate the metric called Forgetting Percentage Points ($FPP$) followed by CL-DETR~\cite{liu2023continual}: $FPP = AP^1-AP^t_{old}$,
where $AP^1$ evaluates for all classes in the first task and $AP^t_{old}$ evaluates for previous classes learned from the first task during current task. The comparison of False Positives ($FP$) shows that CASA effectively overcomes the \textit{background shift} problem~\cite{liu2023augmented} in IOD.
\subsection{Implementation Details}
Our method CASA is based on OWL-ViT, which combines a Vision Transformer (ViT) with a text encoder, allowing the model to understand both images and text prompts, designed for open-world object detection and classification. All experiments are performed using 8 NVIDIA A100 GPUs.

\subsection{Results and Analyses}

\begin{table}[t]
\centering
\caption{CASA results (\%) on COCO 2017 in \emph{two-phase setting} 40+40. The best performance is highlighted in \textbf{bold}. }
\vskip -0.1in
\renewcommand{\arraystretch}{1.0}
\small
\label{tab:40+40}
 \resizebox{0.45\textwidth}{!}{
\begin{tabular}{c|l|c|cccccc}
\toprule
{Scenarios} & \multicolumn{1}{c|}{{Method}}  & {Baseline} & $AP$ & $AP_{.5}$ & $AP_{.75}$ \\ 
\hline
\multirow{1}{*}{80} 
     & Joint & OWL-ViT & 42.1 & 61.8 & 47.1 \\ \hline
\multirow{2}{*}{40} 
     & --- & Def-DETR & 46.5 & 68.6 & 51.2 \\ 
     & --- & OWL-ViT & 46.0 & 65.7 & 50.8 \\ \hline
\multirow{9}{*}{40 + 40} 
     & ERD & UP-DETR & 35.4 & 55.1 & 38.3 \\ 
     & CL-DETR & UP-DETR & 37.0 & 56.2 & 39.1 \\ 
     & LwF & Def-DETR & 23.9 & 41.5 & 25.0 \\ 
     & iCaRL & Def-DETR & 33.4 & 52.0 & 36.0 \\ 
     & ERD & Def-DETR & 36.0 & 55.2 & 38.7 \\ 
     & CL-DETR & Def-DETR & 37.5 & 55.1 & 40.3 \\ 
     & VLM-PL & Def-DETR & 41.7 & 59.9 & 44.2 \\ 
     & CIOD & Def-DETR & 43.0 & 62.1 & 47.1 \\  
     & Ours & OWL-ViT & \textbf{43.2} & \textbf{62.5} & \textbf{47.2} \\ 
\bottomrule
\end{tabular}}
\vskip -0.1in
\end{table}

\begin{table*}[ht]
\centering
\renewcommand{\arraystretch}{1.0}
\caption{CASA results ($AP$/$AP_{.5}$, \%) on COCO 2017 in \emph{multi-phase setting}. The best performance is highlighted in \textbf{bold}.}
\vskip -0.1in
\label{tab:multi_results}
\small
{
\begin{tabular}{l|c|cccc|cc}
\toprule
\multirow{2}{*}{Method} & \multirow{2}{*}{$\mathcal{T}_1$ (1-40)} & \multicolumn{4}{c|}{40+10+10+10+10} & \multicolumn{2}{c}{40+20+20} \\
&  & $\mathcal{T}_2$ (40-50) & $\mathcal{T}_3$ (50-60) & $\mathcal{T}_4$ (60-70) & $\mathcal{T}_5$ (70-80) & $\mathcal{T}_2$ (40-60) & $\mathcal{T}_3$ (60-80) \\
\midrule
    ERD & \multirow{3}{*}{46.5 / 68.6} & 36.4 / 53.9 & 30.8 / 46.7 & 26.2 / 39.9 & 20.7 / 31.8 & 36.7 / 54.6 & 32.4 / 48.6 \\
    VLM-PL &  & 41.7 / 59.3 & 38.5 / 56.4 & 34.7 / 53.6 & 31.4 / 50.8 & 41.7 / 60.4 & 39.7 / 56.5 \\
    CIOD &  & 42.3 / 62.8 & 40.6 / 60.2 & 40.0 / 59.0 & 36.8 / 54.7 & 42.5 / 62.2 & 41.1 / 59.5 \\
\midrule
    Ours & 46.0 / 65.7  & \textbf{45.5} / \textbf{66.4} & \textbf{43.1} / \textbf{62.5} & \textbf{43.2} / \textbf{62.3} & \textbf{41.5} / \textbf{59.7} & \textbf{43.0} / \textbf{62.5} & \textbf{41.6} / \textbf{60.0} \\
\bottomrule
\end{tabular}}
\vskip -0.1in
\end{table*}

\begin{table*}[t]
\centering
\caption{Ablation results in \emph{multi-phase setting}. The ``---'' indicates lacking object detection capability in the first task. 
}
\renewcommand\arraystretch{1}
\small
\label{tab:ab40}
\setlength{\tabcolsep}{3pt}
\vspace{-8pt}
\begin{tabular}{c c c c c|c|c c c|c c c|c c c|c c c}
\toprule
\multirow{2}{*}{Sel} & \multirow{2}{*}{Ada} & \multirow{2}{*}{Ada+Sha} & \multirow{2}{*}{Ref} & \multirow{2}{*}{Ref+Sha} 
& \multicolumn{1}{c|}{40} & \multicolumn{3}{c|}{40+10} & \multicolumn{3}{c|}{40+10+10} & \multicolumn{3}{c|}{40+10+10+10} & \multicolumn{3}{c}{40+10+10+10+10} \\  
 & & & & & All & All & Old & $FPP$ & All & Old & $FPP$ & All & Old & $FPP$ & All & Old & $FPP$ \\  
\hline
$\checkmark$ &   &   &   &   &  0.13 & 0.09 & 0.11 & --- & 0.08 & 0.09 & --- & 0.12 & 0.08 & --- & 0.12 & 0.13 & --- \\
$\checkmark$ & $\checkmark$ &   &   &   & 64.61 & 14.02 & 0.18 & 64.43 & 7.62 & 0.16 & 8.31 & 9.58 & 0.15 & 7.47 & 6.70 & 0.15 & 9.43 \\
$\checkmark$ &   &   & $\checkmark$ &   &  0.16  & 0.12 & 0.13 & --- & 0.07 & 0.08 & --- & 0.11 & 0.08 & --- & 0.12 & 0.13 & --- \\
$\checkmark$ & $\checkmark$ &   & $\checkmark$ &   & \textbf{65.67}& 14.04 & 0.10 & 65.57 & 7.99 & 0.13 & 13.91 & 9.56 & 0.09 & 7.90 & 6.74 & 0.12 & 9.44 \\
$\checkmark$ &   & $\checkmark$ &   &   & 64.61 & 65.21 & 64.98 & \textbf{-0.22} & 61.11 & 65.43 & \textbf{-0.48} & 60.59 & 61.59 & \textbf{-0.29} & 57.75 & 60.88 & \textbf{-0.29} \\
$\checkmark$ &   &   &   & $\checkmark$ & 0.15 & 0.12 & 0.15 & --- & 0.15 & 0.15 & --- & 0.17 & 0.16 & --- & 0.15 & 0.16 & --- \\
$\checkmark$ &   & $\checkmark$ &   & $\checkmark$ &\textbf{65.67} &\textbf{66.33} & \textbf{65.67} & 0.03 &\textbf{62.54} & \textbf{66.30} & -0.02 & \textbf{62.32} &\textbf{62.56} & 0.07 & \textbf{59.67} &\textbf{62.25} & 0.07 \\
\bottomrule
\end{tabular} 
\end{table*}

\subsubsection{Two-phase setting}
We randomly divide the 80 classes of the COCO dataset into two experimental settings: 70+10 and 40+40. In Tab~\ref{tab:70+10} and Tab~\ref{tab:40+40} we compare the performance of our method, CASA, with other IOD methods in terms of the metrics $AP$, $AP_{.5}$ and $AP_{.75}$ in both experimental settings. Additionally, in the 70+10 setting, we compare the difference in $FPP$ between our method and the previously best IOD method, demonstrating that our method not only prevents forgetting but also achieves better performance than the previous task. As for False Positives ($FP$), in the 70+10 setting CASA has 31052 errors, which demonstrate an clear advantage, reducing at least 5000 errors than other methods. This indicates that our method effectively overcomes the issue of \textit{background shift} compared to other IOD methods. Both in 70+10 and 40+40 settings, CASA consistently outperforms the state-of-the-art, better than CIOD and other IOD methods. In the 70+10 setting, our method achieves a 1.5\% improvement in $AP_{.5}$ and a 1.9\% improvement in $FPP$ over the current best methods, which is significantly more evident than the performance gains in the 40+40 setting. This is because CASA leverages class-agnostic shared attribute information for incremental object detection. The more classes learned in the first phase, the richer the preserved shared attribute information, which benefits the subsequent phases of incremental learning.

\begin{figure}[tbp]
\vskip -0.2in
\begin{center}
\centerline{\includegraphics[width=\columnwidth]{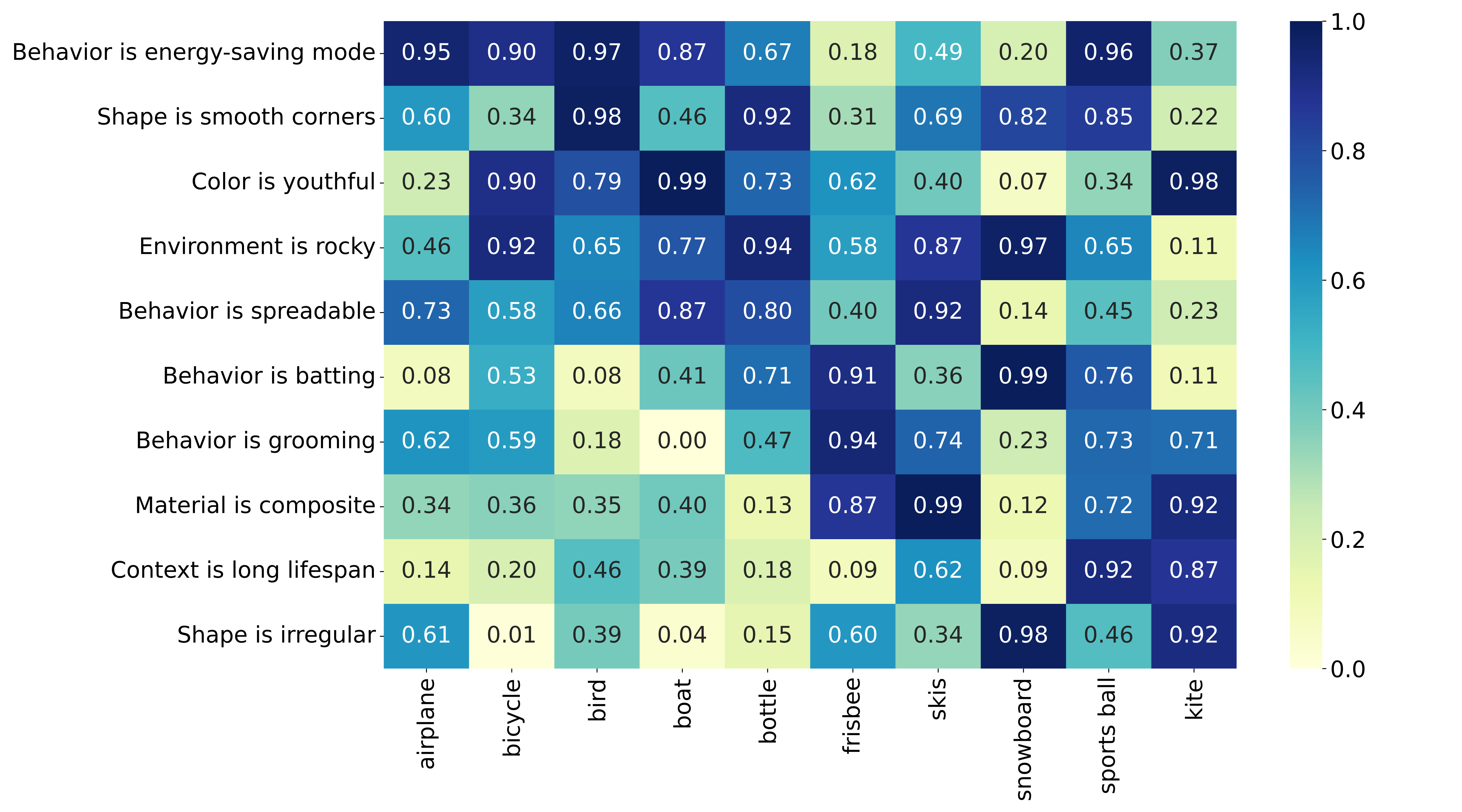}}
\vskip -0.1in
\caption{
Attribute scores in two-phase setting.The first five classes belong to the initial phase, while the latter five classes are part of the second phase.
}
\vskip -0.2in
\label{fig:att}
\end{center}
\vskip -0.2in
\end{figure}

\subsubsection{Multi-phase setting}
We conduct experiments in 40+20+20 and 40+10+10+10+10 settings respectively. As before, the categories for each phase are randomly assigned. Tab~\ref{tab:multi_results} records the $AP$ and $AP_{.5}$ at each step. We observe that CASA exhibits the best performance among IOD methods in both multiple-phase settings, though the performance in the first phase is inferior to that of the method based on Deformable DETR. Notably, in the 40+10+10+10 setting, CASA demonstrates strong continuous learning capabilities across multiple steps, with $AP$ and $AP_{.5}$ significantly improved by 4.7\% and 5\%, respectively, compared to the current state-of-the-art method. This effectively indicates that our method can be applied to real-world scenarios for continuous learning.

\subsection{Class-agnostic Shared Attributes}

Figure~\ref{fig:att} shows the attribute scores in the two-phase setting. We observe that the first task exhibits high scores across the first five attributes, and the last five classes in the second task show high scores for some of these attributes, indicating that certain attributes are shared across different tasks. In the second task, the scores for the subsequent five attributes are notably higher. Some of these attributes had lower scores in the first task, suggesting that the second task would select a new set of shared attributes.

Besides, The third part of supplementary materials IV provides an example of the process of increasing shared attributes.

\subsection{Ablation Experiments}
We perform ablation experiments on three modules: attributes selection and the assignment matrix $A^t$ updating, attributes adaption and attributes refinement. Table~\ref{tab:ab70} and Table~\ref{tab:ab40} show that all three components are essential. Detailed explanation is in supplementary materials IV.

\begin{table}[t]
\centering
\caption{Ablation results in \emph{two-phase setting} 70+10.}
\vskip -0.1in
\small
\label{tab:ab70}
\setlength{\tabcolsep}{1.5pt}
\begin{tabular}{c c c c c|c|c c c}
\toprule
\multirow{2}{*}{Sel} & \multirow{2}{*}{Ada} & \multirow{2}{*}{Ada+Sha} & \multirow{2}{*}{Ref} & \multirow{2}{*}{Ref+Sha} & \multicolumn{1}{c|}{70} & \multirow{3}{*} & 70+10 & \\ 
 &  &  &  &  & All & All & Old& $FPP$\\ 
\hline
$\checkmark$ &   &   &   &   &  0.18 & 0.17 & 0.18 & --- \\
$\checkmark$ & $\checkmark$ &   &   &   &  62.99 & 0.15 & 0.16 & 62.83 \\
$\checkmark$ &   &   & $\checkmark$ &   &  0.13 & 0.15 & 0.16 & --- \\
$\checkmark$ & $\checkmark$ &   & $\checkmark$ &   &  \textbf{63.43} & 6.75 & 0.11 & 63.32 \\
$\checkmark$ &   & $\checkmark$ &   &   &  62.99 & 59.71 & 62.42 & 0.57 \\
$\checkmark$ &   &   &   & $\checkmark$ &  0.13 & 0.12 & 0.12 & --- \\
$\checkmark$ &   & $\checkmark$ &   & $\checkmark$ &  \textbf{63.43} & \textbf{61.00} & \textbf{63.44} & \textbf{-0.01} \\
\bottomrule
\end{tabular} 
\label{tab:evaluation}
\end{table}

\begin{figure}[tbp]
\begin{center}
\centerline{\includegraphics[width=0.75\columnwidth]{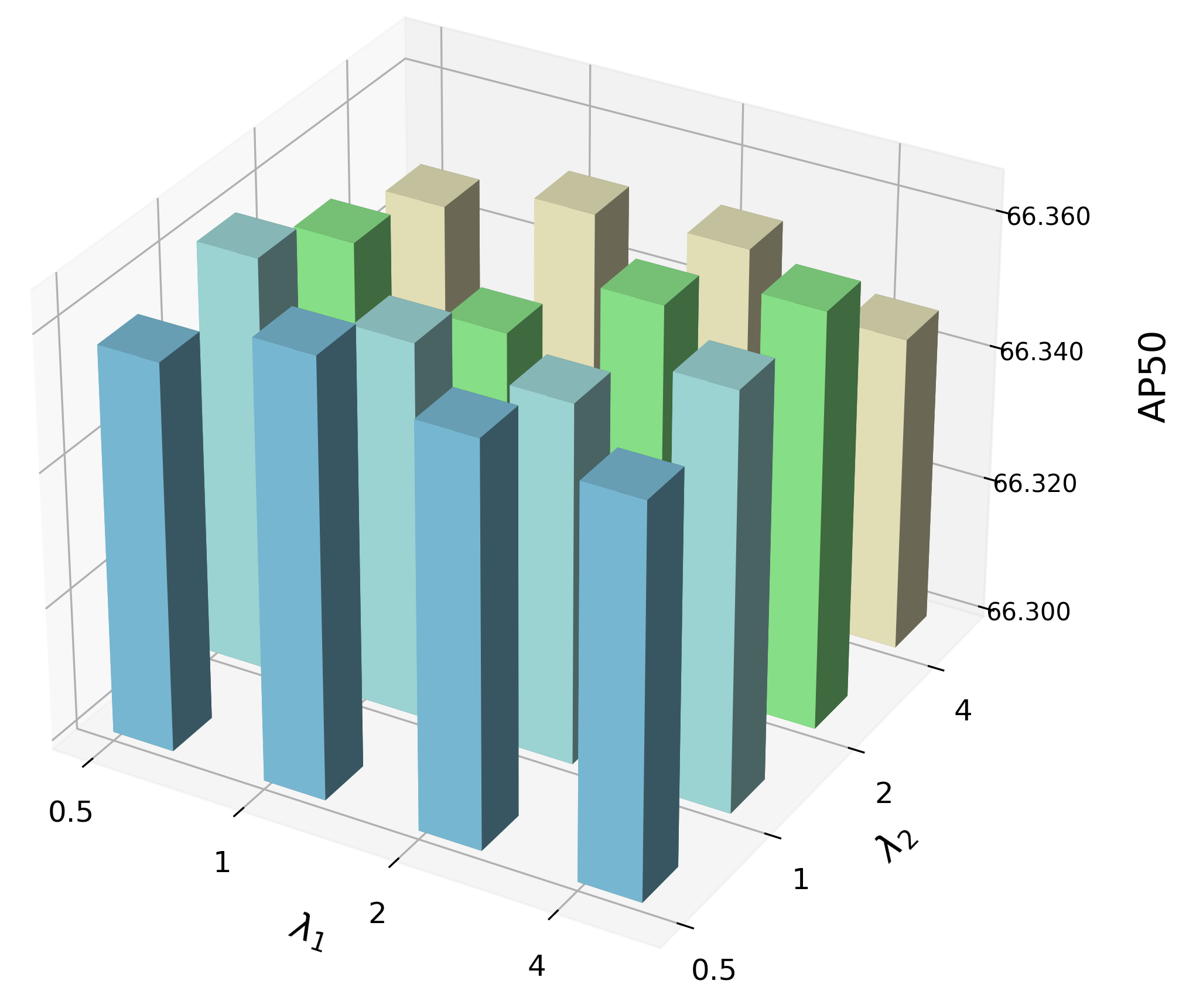}}
\vskip -0.1in
\caption{
Impact of the hyperparameters $\lambda_1$ and $\lambda_2$
}
\vskip -0.2in
\label{fig:lamda}
\end{center}
\end{figure}

Specifically, we experiment with not sharing attribute across the three modules and not applying the loss function that enforces the consistency of attribute embedding between phases. Results indicate that sharing attribute information and applying the loss function that controls phase-to-phase consistency lead to the best performance in incremental object detection.

\subsection{Effect of Hyperparameters and Visualizations}

Figure~\ref{fig:lamda} demonstrates the variation in $AP_{.5}$ values under the 70+10 setting when $\lambda_1$ and $\lambda_2$(\textbf{described in supplemental materials III}) are set to 0.5, 1, 2, and 4. We observed that the change in $AP_{.5}$ is less than 0.02\%, indicating minimal impact($\lambda$ fixed  at 0.01). The visualization results of IOD are shown in the figure~\ref{fig:viz}. Detailed explanation is also in supplementary materials V.

\begin{figure}[tbp]
  \centering
  \begin{subfigure}[b]{0.24\textwidth}
    \includegraphics[width=\textwidth, height=0.7\textwidth]{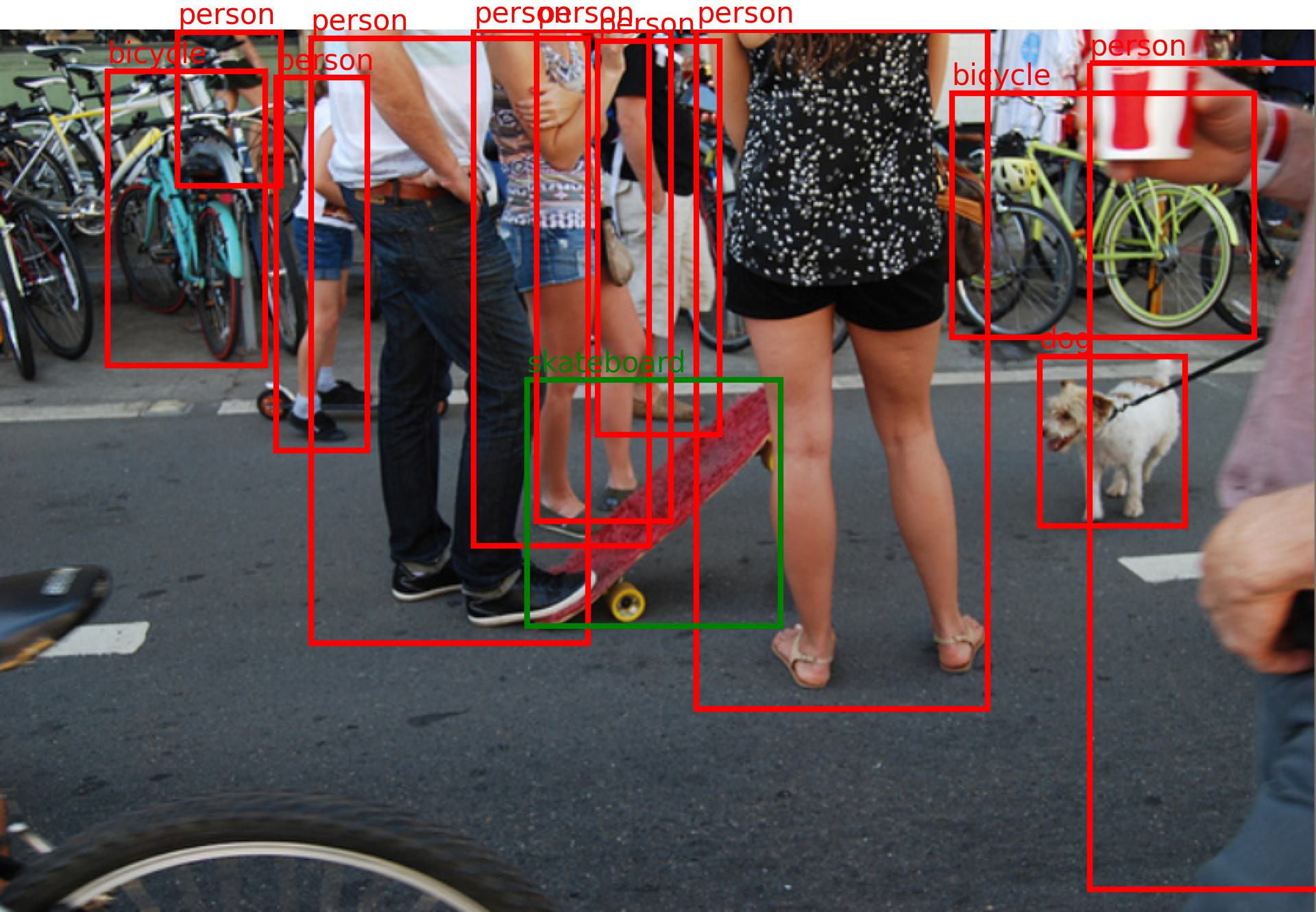}
    \includegraphics[width=\textwidth, height=0.7\textwidth]{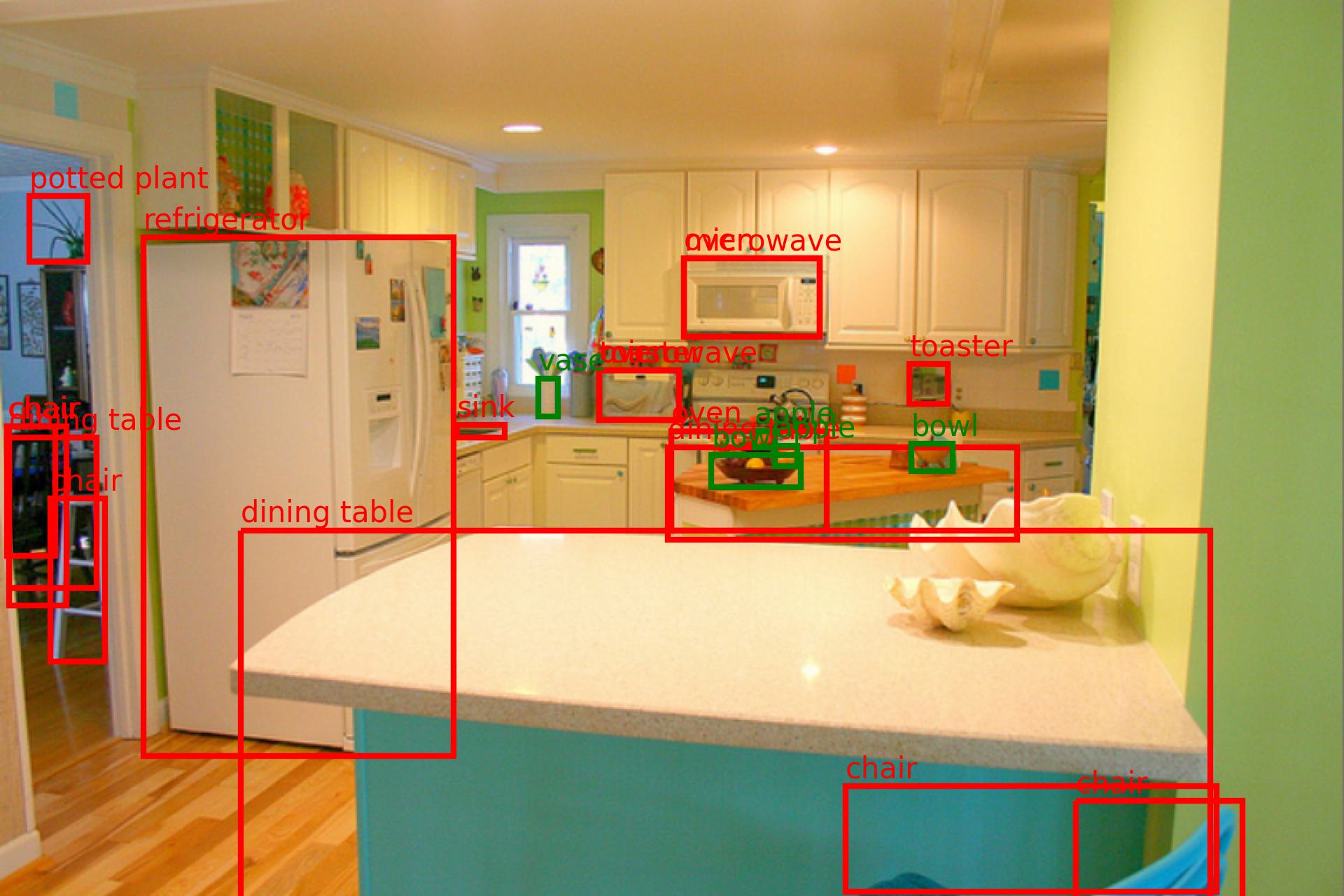}
  \end{subfigure}
  \begin{subfigure}[b]{0.24\textwidth}
    \includegraphics[width=1.02\textwidth, height=0.7\textwidth]{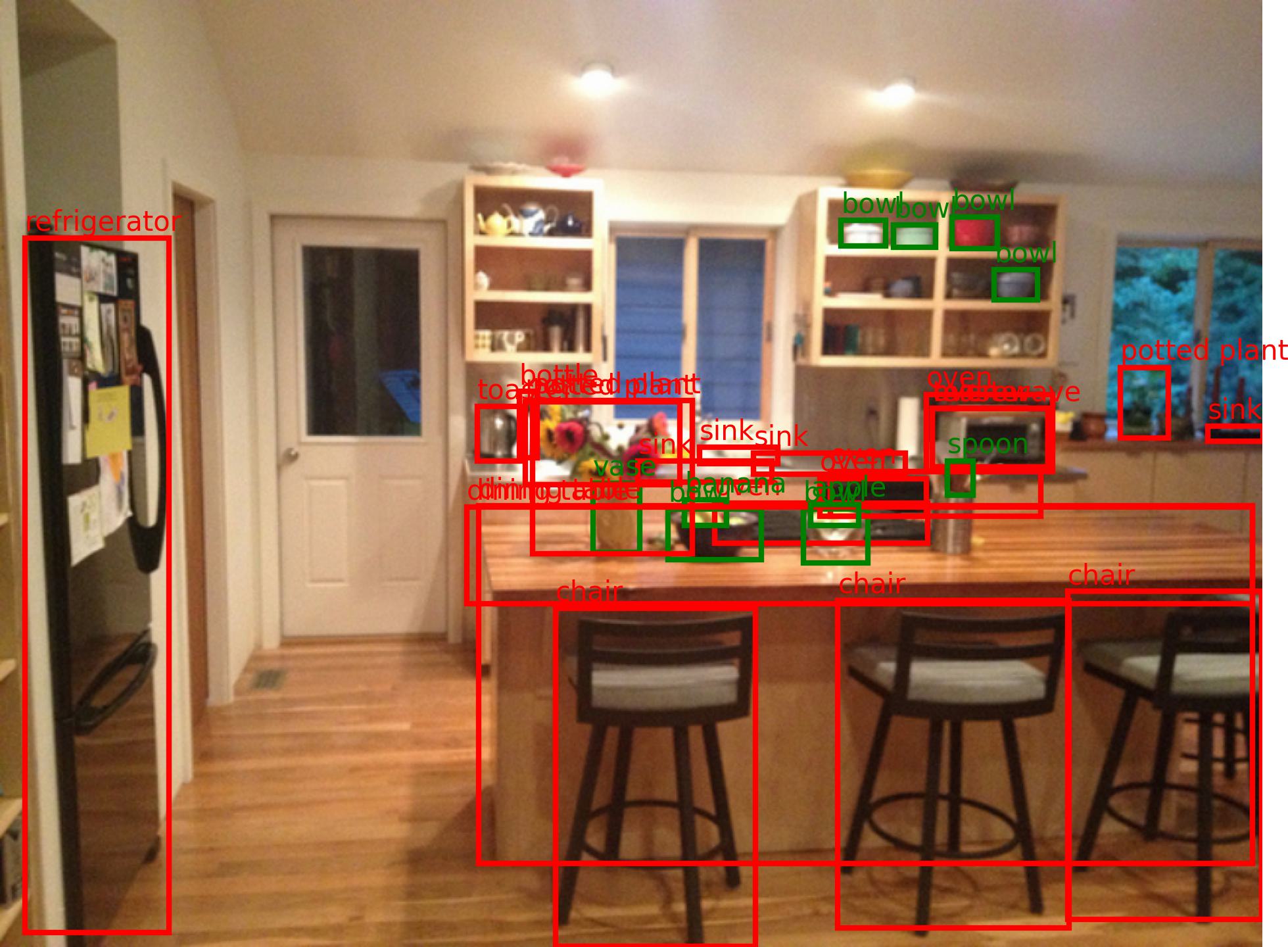}
    \includegraphics[width=\textwidth, height=0.7\textwidth]{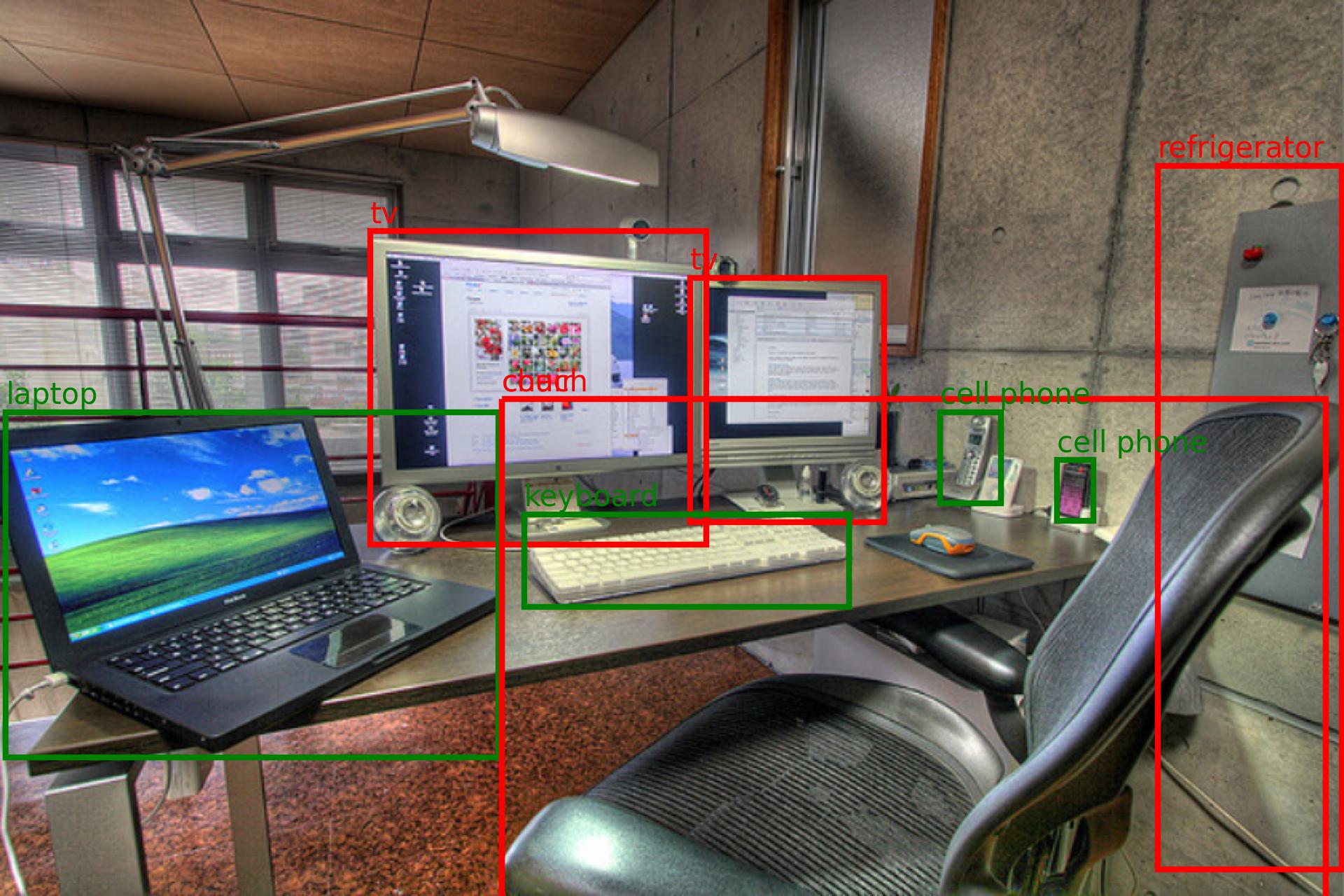}
  \end{subfigure}
  \caption{Visualization results for 40+40 setting. The red boxes show object classes learned in the previous phase, while the green boxes represent those learned in the current phase.}
  \label{fig:viz}
  \vspace{-10pt}
\end{figure}
\section{Conclusions}
In this paper, we propose a novel method for IOD, CASA, which effectively addresses the challenges of \textit{background shift} by leveraging vision-language foundation models. By constructing a Class-Agnostic Shared Attribute base, CASA captures and retains common semantic attributes across incremental classes. Our method preserves parameters of the pre-trained OWL-ViT model while incorporating parameter-efficient fine-tuning, which only adds 0.7\% of parameter storage. Extensive experiments demonstrate that CASA achieves state-of-the-art performance in both two-phase and multi-phase incremental learning scenarios. 

\bibliographystyle{IEEEbib}
\bibliography{icme2025references}

\vspace{12pt}

\newpage

\twocolumn[
  \begin{@twocolumnfalse}
    \begin{center}
      \Huge\bfseries Supplemental Materials
      \vspace{1.7em} 
    \end{center}
  \end{@twocolumnfalse}
]

\setcounter{section}{0}
\section{Details of attributes}
\subsection{Shared Attributes Base}
We use ten types of attributes, which are Color(such as red, yellow), Shape, Texture, Size, Context, Features, Appearance, Behavior, Environment, and Material. The attributeas base is generated only once. For each task, we selects a set of representive attributes using for detection from this attributes base according to their scores. For different tasks, the attributes used for detection are expanded, and shared among previous and new classes.
\subsection{Filtering out Unused Attributes}
The initial Shared Attributes Base is generated with LLM. Some attributes are representative and effectively distinguish between different classes, but others, such as “shape is irregular,” do not significantly represent classes. We remove such attributes for two main reasons: on the one hand, this greatly reduces computational overhead; on the other hand, these attributes tend to have similar matching scores, which impact the subsequent process and reduce the accuracy of detection results. 

When a new task arrives, we scores every attribute in the attribute base for new classes (including previously removed attributes). Attributes with high scores are retained, which may include previously selected attributes or previously removed ones. If an attribute is removed in a prior task but selected for a new task, our method appends these attributes to the end of the previous task's remained attributes used for detection. Thus, the new set of attributes used for detection consists of attributes used in previous tasks and those newly selected in the current task.
\subsection{Process of Increasing Shared Attributes}
Here, we illustrate the process of increasing shared attributes using the 70+10 experimental setting as an example. In the first phase, we assume that each category is represented by 25 attribute information. In the first task, CASA only select 1314 attributes out of 2895 possible attributes for 70 categories, indicating that many attributes are shared among different categories within the first phase. In the second phase, CASA learns 10 additional categories, and in practice only 155 new attributes are added. This is partly due to some attributes being shared among these 10 categories, and more importantly, the attributes of these 10 categories are already used among the previous 70 categories, achieving efficient attribute sharing between the previous and current tasks. 

\section{Implementation Details Before Inference}

\subsection{Attributes Adaption}
\setcounter{equation}{0}
After selecting the attribute embedding $\hat{E}_a^t$ based on $A^{t-1:t}$ for the current task, we also need to adapt the attribute embedding $\hat{E}_a^t$ to eliminate barriers between visual and textual information, because the textual attribute information generated from the text and the visual information are orthogonal in space, belonging to different domains. For each category in the current task, we take \( M \) samples and calculate the visual mean embedding $\bar{E}^v$ for these \( M \) samples:
\begin{equation}
    \bar{E}^v(c) = \frac{1}
    {M}\sum_{i=0}^{M}\mathbf{e}^v_i.
\end{equation}

Our adapting strategy for $\hat{E}_a^t$ aims to align the transpose of $A^{t-1:t}$ with $\hat{E}_a^t$. Additionally, to achieve incremental object detection, in Task $T_t$ we need to 
ensure that the first $Q$ rows of $\hat{E}_a^t$, $\hat{E}_a^t[\ :Q, :]$, is as consistent as possible with $\hat{E}_a^{t-1}$, where $Q$ equals the number of rows in $\hat{E}_a^{t-1}$. This can mitigate the forgetting of previously learned classes.The loss function \( \mathcal{L}_{ada} \) can be formulated as:
\begin{equation}
\begin{aligned}
\mathcal{L}_{ada} = &\ \mathcal{L}_{MSE}\left(\bar{E}^v\  , {A^{t-1:t}}^\top \otimes \hat{E}_a^t\right) \\
&+ \lambda_1\mathcal{L}_{MSE}\left(\hat{E}_a^t[\ :Q, :], \hat{E}_a^{t-1}\right),
\end{aligned}
\end{equation}
where $\lambda_1$ is a tunable hyperparameter. In this way, we adapt the attribute embedding $\hat{E}_a^t$, eliminating barriers between visual and textual information, which can be used for future refinement.
\subsection{Attributes Refinement}
In Task $T_t$, after adapting the attribute information, we perform refining on a small scale to better apply the assignment matrix $A^t$ and the attribute embedding $\hat{E}_a^t$ to the subsequent inference. Similar to Equation 1 in the main text, we first compute the similarity vector $\hat{S}_t$ between the attribute embedding $\hat{E}_a^t$ and visual embedding $\mathbf{e}^v$ in the current task. Next, with the assignment matrix $A^{t-1:t}$, logits corresponding to these $C_t$ categories can be computed using Equation 2 in the main text. Unlike the previous process, at this stage, we need to keep the assignment matrix $A^t$ unchanged and update $\hat{E}_a^t$. We calculate the BCE loss between the probabilities $P$ of these categories and the targets $U$, which are a list of labels corresponding to the target categories in the image.

To achieve incremental object detection, we also need to ensure that the first $Q$ rows of the current task's $\hat{E}_a^t$, $\hat{E}_a^t[\ :Q, :]$, remains consistent with the $\hat{E}_a^{t-1}$ saved from the previous task. Therefore, an additional BCE loss between them is added. The loss function at this stage can be expressed as:
\begin{equation}
\begin{aligned}
\mathcal{L}_{ref} = &\ \mathcal{L}_{BCE}\left(P, U\right) + \lambda_2\mathcal{L}_{MSE}\left(\hat{E}_a^t[\ :Q, :], \hat{E}_a^{t-1}\right),
\end{aligned}
\end{equation}where $\lambda_2$ is also a tunable hyperparameter. After refining the attribute embedding, the $\hat{E}_a^t$ and the $A^t$ can be used in the following inference stage.

\section{Hyperparameters Setting and parameter storage}
\subsection{Hyperparameters Setting}
During training, we set the number of epochs \([1, 10, 100]\), and the learning rate \([1\text{e-6}, 5\text{e-6}, 1\text{e-5}, 5\text{e-5}, 1\text{e-4}]\), searching for the best pair of epochs and learning rate. We fix the hyperparameter \(\lambda\) at 0.01 used for the regularization of the assignment matrix, while the optimal hyperparameters \(\lambda_1\) and \(\lambda_2\), which control the consistency of attribute embeddings between the current task and the previous task, are adjustable for different experimental settings.
\subsection{Parameter Storage}
It is noteworthy that we do not retrain the OWL-ViT model, instead we achieve efficient fine-tuning of the pre-trained model. The parameter storage number of OWL-ViT is $4.31\times10^8$, while our fine-tuned method increases the parameter storage number to $4.34\times10^8$, representing an increase of less than \textbf{0.7\%}. 

\section{Ablation Experiments' explanation}
As shown in Table V and Table IV in the main text, results show that all three components(attributes selection and the assignment matrix $A^t$ updating, attributes adaption and attributes refinement) are essential. The first and second modules are indispensable, as the absence of either module leads to the model's inability to perform effective detection. The inclusion of the refining module enhances the model's detection results. Specifically, although in certain cases, methods that do not involve refinement may result in a slightly lower $FPP$, the detection metrics for each task, such as $AP$, $AP_{.5}$ and $AP_{.75}$, will show significant discrepancies compared to methods that do include refinement, and these discrepancies will continue to widen over time. Therefore, refinement remains necessary.

\section{Visualization details}
Figure 4 in the main text presents the visualization results under the 40+40 experimental setup on the COCO dataset. The red boxes represent the object classes learned by the model in the previous phase, while the green boxes indicate the object classes learned by the model in the current phase. From the visualization results, it can be observed that our method, CASA, demonstrates excellent detection performance for both the current and previously learned classes, effectively achieving incremental learning and overcoming \textit{background shift}.
\end{document}